# Multi-Agent Path Planning Using Deep Reinforcement Learning

Mert Çetinkaya

**Abstract.** In this paper a deep reinforcement based multi-agent path planning approach is introduced. The experiments are realized in a simulation environment and in this environment different multi-agent path planning problems are produced. The produced problems are actually similar to a vehicle routing problem and they are solved using multi-agent deep reinforcement learning. In the simulation environment, the model is trained on different consecutive problems in this way and, as the time passes, it is observed that the model's performance to solve a problem increases. Always the same simulation environment is used and only the location of target points for the agents to visit is changed. This contributes the model to learn its environment and the right attitude against a problem as the episodes pass. At the end, a model who has already learned a lot to solve a path planning or routing problem in this environment is obtained and this model can already find a nice and instant solution to a given unseen problem even without any training. In routing problems, standard mathematical modeling or heuristics seem to suffer from high computational time to find the solution and it is also difficult and critical to find an instant solution. In this paper a new solution method against these points is proposed and its efficiency is proven experimentally.

## 1 Introduction

Path planning or vehicle routing is challenging and this is an NP-hard combinatorial optimization problem. It has lots of different variations and these variations are sourced by some constraints like capacity or time constraints [1, 2]. Because of its very wide use in real word, these problems are quite popular and they get highly attention in academy or real world and different solutions are offered against these problems. The offered solutions are mostly based on operations research based classical mathematical modeling or some heuristics like genetic algorithms, simulated annealing or tabu search. These heuristics are actually produced against the fact that usage of classical operations research requires a lot of time to solve this NP-hard problem and thanks to the heuristics this great time requirement is lessened to a certain extent. However, even using these heuristics, it is still not easy and time requirement is still an important concern.

In this paper, this issue is addressed and a faster approach is proposed to solve a path planning problem. To realize the experiments, a simulation environment is imported from github [3] and is modified according to the needs of desired problem. Then, a random environment is produced with some unusable paths that represent the walls in some parts and a common start and end point in the center like the depot in a vehicle routing problem. The details of the environment will be explained in Section 3. In this fixed environment, a deep reinforcement model is trained for different problems. The problems differ by the places of the target points that the agents must visit and in this way, the produced problem is quite similar to a vehicle routing problem. Each target point (or landmark) must be visited at least once by an agent and at the end each agent must return back to the start point with the aim of minimizing sum of distances covered. A deep reinforcement learning model is trained from scratch to solve the produced different problems explained in this way. As the time and problems are passed, the model's performance seems to increase against the problems, because it starts to learn its environment and directs the agents in a more intelligent way. After enough training, the model becomes capable of finding a nice solution against a totally new problem, a short training can also be benefitted to improve the solution and this is the main aim of this paper. The time requirement to solve a combinatorial optimization problem is totally removed or it is decreased dramatically and this is done using a pre-trained model.

About the remaining parts of this paper, in Section 2 related work in this working area is presented, in Section 3 the simulation environment is shown, in Section 4 reinforcement learning and deep reinforcement learning is explained, in Section 5 experimental study is presented and in Section 6 conclusion is given.

## 2    Related Work

As it is previously explained, path planning or vehicle routing is a popular working area. In some studies path planning of only one agent is concerned whilst in some studies path planning for multiple agents is concerned. Also about the solution method, more classical approaches like integer programming based mathematical modeling or heuristic optimization can be used and they are quite popular. However, usage of machine learning or deep learning has also drawn lots of attention since last years and combination of these artificial intelligence based approaches with mathematical programming or heuristic optimization is also preferred.

In their study, Xizhi et al. [4] and Wang et al. [5] worked on path planning problem and made use of A* algorithm. Li et al. [6] also worked on this problem and in their study, they combined A* algorithm with Genetic Algorithms to obtain better results. Shi et al. [7] used ant colony optimization for global path planning and artificial potential field method for local planning. These are all different studies for path planning of one agent.

Looking at multi-agent path planning problem, Rabbani et al. [8] worked on multi-compartment vehicle routing problem and they also evaluated greenhouse emissions. They presented the related mathematical model and used genetic algorithm and simulated annealing separately and in a hybrid fashion. Kwansang and Chaiwuttisak [9] worked on a vehicle routing problem with some capacity constraint and they used ant colony optimization and improved the solution using 2-Opt and one-move heuristics.

There also exists some studies where reinforcement learning or deep reinforcement learning is used for multi-agent path planning. Lin et al. [10] used deep reinforcement learning to solve electric vehicle routing problem. They trained their model using policy gradient. Similarly, Zhang et al. [11] also used deep reinforcement learning to solve multi-vehicle routing problem. They used an encoder-decoder framework. Qin et al. [12] also worked on this problem and they formulated a mixed-integer linear programming model to solve it. They also combined the model with reinforcement learning and deep learning for improvement. Aljohani et al. [13] worked on routing optimization of electric vehicles. They used deep reinforcement learning and Markov chains and considered driver patterns, environmental effects and road conditions with the aim of minimization of energy utilization.

## 3    Simulation Environment

The simulation environment is imported from github as it is previously explained and it is modified to meet the requirements of the desired path planning problem. A 7x7 simulation environment formed by grids is used and the center point is thought as the fixed start and finish point where the agents start and finish their movement. The walls are also added and they are shown with stop sign. Their positions are also fixed. In this fixed simulation environment the deep learning model is trained using different problems where 5 landmarks (target points) shown by the red flags are scattered randomly. The experiments are always realized with 5 landmarks and 3 agents and the aim is finding the paths of 3 agents that minimize total traveled distance while covering each landmark at least one time. In Fig. 1, the agent and a random environment (or problem) are presented as example. The details of the experimental study are explained in Section 5.

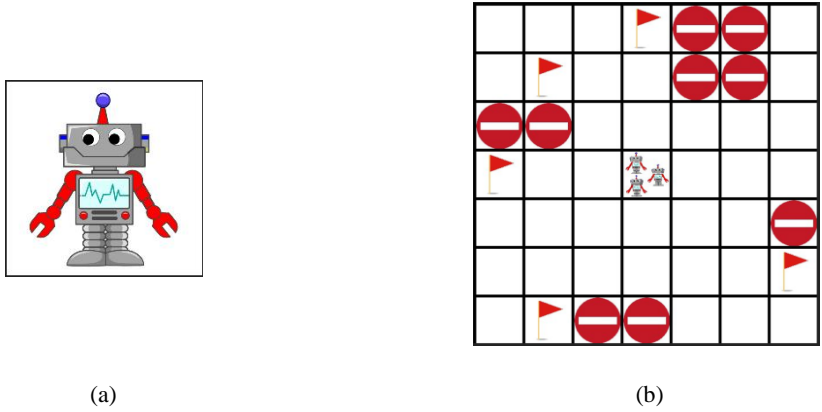

|         |         |
|:-------:|:-------:|
|   (a)   |   (b)   |

**Fig. 1.** Images for presenting the simulation environment and its components. (a) The agent. (b) A random simulation environment.

## 4 Reinforcement Learning and Deep Reinforcement Learning

In a simple single-agent reinforcement learning, there becomes an agent and it interacts with its environment through its actions. At each time $t$, the agent observes its current state $s$ and chooses an action $a$ according to a policy and it receives a reward $r$. For a desired behavior, it receives a high reward, but for an undesired behavior a low reward value is given to the agent. In this way, agent is trained and it learns to choose the best action for a state after enough training. In reinforcement learning, a quite popular and widely used algorithm is Q-Learning. In Q-Learning, $Q$ function is used and updated to determine the behavior of the agent. Based on its experience, it is updated according to Eq. 1 by the agent. Here $\alpha$ is learning rate and $\gamma$ is discount rate [14].

$$Q_{t+1}(s,a) = Q_t(s,a) + \alpha_t * [r_t + \gamma * \max(Q_{t+1}(s,a)) - Q_t(s,a)] \quad (1)$$

In the above equation, standard Q-Learning is introduced. However this Q-Learning is quite limited, because it is a tabular approach and when there are lots of states that cannot be kept in a table, this method remains insufficient. The tabular Q-learning can only work with finite or less number of states. Here, the solution becomes Deep Q-Learning (Deep Q-Networks) (DQN) [15]. Instead of a table, a network is used in this approach and its weights are updated using the old experiences whose special name is replay memory. State information is used as input of the network and in this way, limitation problem for state number is overcome. As more improved DQN approaches, Double DQN [16] or Dueling Double DQN [17] can be shown. In this paper, simply DQN is preferred.

## 5 Experimental Study

In the experimental study, the simulation environment given in Fig. 1(b) is used and landmark positions are changed in this environment. Each different dispersion of landmarks represent a different problem and our model is trained on different and random problems. The model is trained on one problem during 30 episodes and an episode consists of maximum 50 steps. An agent can move one unit towards left, right, up or down in a step and in our system, the agents always move together and take a step at the same time. A step is taken when all of the agents realize one of the four movements at the same time. In this way, the model is trained on 20 different randomly produced problems.

In a multi-agent reinforcement learning problem, there can be 2 strategies. One of them is training one model like a 'meta-agent' that takes the information of all agents and collects them as input and directs all the agents according to the output of the input. The other is training independent models, each of them belongs to one agent. In our case, it is necessary to see the system and all the agents as a whole, because the aim is minimizing the total distance traveled by all the agents while visiting all the landmarks. That's why, the first approach which is training one model is preferred.

For this model, the input information (state representation) is position information of agents, position information of landmarks and visiting information of landmarks as a binary variable. Because there exists 3 agents and 5 landmarks, the length of input vector becomes 13 (3 for agents' position, 5 for landmarks' position and 5 for landmarks' visiting information). The position information for an agent or landmark is represented using only one number for simplification. The squares in the 7x7 environment are enumerated in [0, 48] interval.

The output of the model is the $Q$ scores for 4 movements of each agent. Because there exists 3 agents, the length of output vector becomes 12. The 4 movements are moving one unit towards left, right, up or down and the one with the highest score is chosen by each agent. About the reward and termination policy of an episode, when all the landmarks are visited, reward becomes 0 and an episode is finished. Otherwise, for each unvisited landmark, Manhattan distance ($l_1 -$ norm) is calculated between each agent and that landmark and minimum of the distance values is taken. These minimum distance values are calculated for each unvisited landmarks and negative of their sum becomes the reward value for that step.

Lastly, about the network architecture, the DQN architecture described by Mnih et al. [15] is used with some modifications. The convolutional layers are removed and a 3-layer fully-connected neural network whose input is state representation is used. In the first 2 layers 512 rectifier units are used in each layer and these layers are followed by the linear output layer which has 12 units. For some more details about hyper-parameters, Adam optimization algorithm is used with 0.001 learning rate and 0.95 discount factor. A finite and FIFO based replay memory is also used during training.

Following this architecture and training principle, the training is carried out for 20 randomly produced problems and for each problem training is continued during 30 episodes (600-episode training in total). At the end of this training the plots given in Fig. 2 and Fig. 3 are obtained for reward and step values for the episodes. The training and all the other experiments are realized on an ordinary laptop computer and the training takes approximately 3 hours.

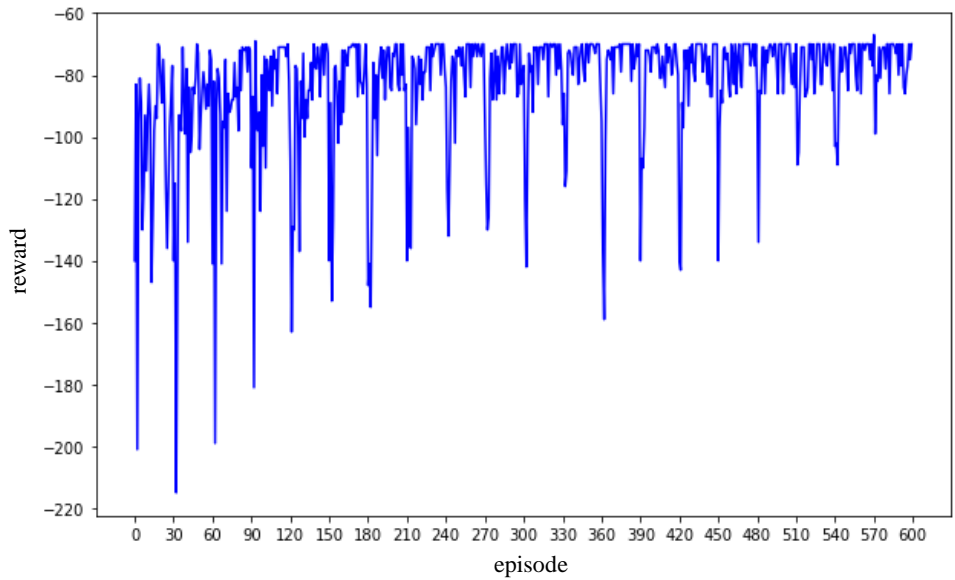

**Fig. 2.** Reward values for each episode during training.

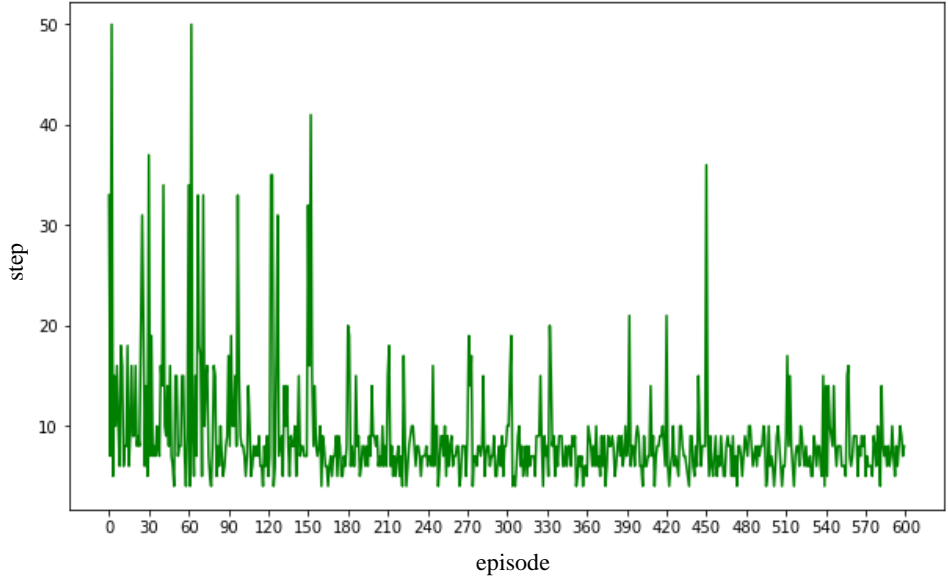

**Fig. 3.** Step numbers for each episode during training.

Regarding Fig. 2 and Fig. 3, it is seen that as the episodes pass and training continues, the model starts to learn the environment and it can produce a nice solution to a given problem. There also exists some unwanted peaks, and the source for them can be the stochastic nature of the optimization algorithm in training process and the complicated inner structure of deep reinforcement learning. However, regarding the general pattern, it is obvious that there is an efficient training and the length of peaks also decreases as the training continues. If training is continued even more after 600 episodes, it seems that a better model will be obtained.

As a general training principle, the rewards and step number values are determined with regards to agents' movement until all the landmarks are visited. Then, some very simple heuristics are used to finish an agent's movement after the last landmark it visits. For example, if the center point is at the right of the agent, it is directed towards right. These kind of heuristics are also used during training of the model. For example, an agent is directed towards left if there is an unvisited and close landmark at the left. The influence of these heuristics to visit landmarks are larger at the beginning to accelerate the training process and as the model is trained their influence gets lessened and limited.

For example, in Fig. 4, 2 different unseen problems and the solution of produced model at the end of 600-episode of training is given. The figures show that the produced model can find nice solution on unseen problems for route planning and can make a nice route planning. The arrows on the route show movement direction of the agents.

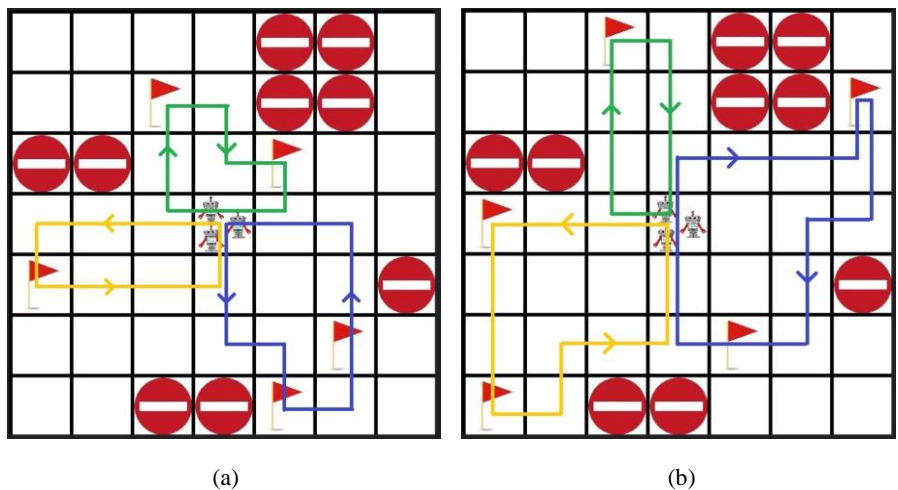

(a)          (b)

**Fig. 4.** Solutions for two different problems. (a) First problem. (b) Second problem.

To improve even more the produced solution for any unseen problem, a short training can be useful on the problem. In this way, the model can learn better the dynamics of the given environment and a better solution can be found.

# 6   Conclusion

In this paper, the problem of high time requirement in path or route planning problems is targeted. A method that gives instant and satisfying solutions is presented for these problems and it can be very useful for systems that work in dynamic environments and require instant solutions. As an example, meal delivery or virtual market services can be shown. For these kind of services, it is very important to quickly meet the changing demands in a dynamic environment. Usage of reinforcement learning or deep reinforcement learning in these kind of delivery problems is also an interesting field of study as a variant of route planning [18, 19].

To realize the study, a simulation environment is produced and the experimental study is carried out on this environment. Because of the high computational requirement, a limited and small simulation environment is used and from this point of view this is actually a proof-of-concept study, but its results are quite promising. The experimental study is realized on an ordinary laptop computer and to realize a more realistic study, a larger simulation environment or a real world problem must be chosen and (much) longer model training must be preferred on a much more powerful computer or a server.

Lastly, to improve model results, more intelligent deep reinforcement learning approaches like double DQN or dueling double DQN and some mathematical modeling or heuristic optimization can also be benefitted and different approaches can be combined.